\documentclass[sigconf,nonacm]{acmart}

\usepackage{amsmath}
\usepackage{graphicx}
\AtBeginDocument{\raggedbottom}

\newcommand{\NMissions}{255}
\newcommand{\NSuccess}{244}
\newcommand{\NFail}{11}
\newcommand{\SuccessRate}{95.7}

\newcommand{\NModels}{3}
\newcommand{\NScenarios}{4}

\newcommand{\NScalingMissions}{150}
\newcommand{\MaxSwarm}{5}
\newcommand{\FlightGain}{28.1}
\newcommand{\TotalGain}{18.9}
\newcommand{\LawFixed}{332}
\newcommand{\LawDivisible}{174}
\newcommand{\LawP}{0.34}
\newcommand{\LawPLow}{0.29}
\newcommand{\LawPHigh}{0.40}
\newcommand{\SpeedLimit}{1.52}
\newcommand{\SpeedLimitLow}{1.40}
\newcommand{\SpeedLimitHigh}{1.67}

\newcommand{\PowerAlpha}{0.20}
\newcommand{\AicGap}{0.8}
\newcommand{\AgentIntercept}{140}
\newcommand{\AgentSlope}{6.7}
\newcommand{\AgentSlopeLow}{0.8}
\newcommand{\AgentSlopeHigh}{12.0}

\newcommand{\SecondUavShare}{50}
\newcommand{\MaxSwarmShare}{80}
\newcommand{\FlightOne}{505}

\newcommand{\TotalOne}{657}

\newcommand{\FlightTwo}{426}

\newcommand{\FlightThree}{393}

\newcommand{\FlightFive}{363}

\newcommand{\TotalFive}{533}

\newcommand{\NScalingLocated}{145}
\newcommand{\NCompared}{135}
\newcommand{\NComparedLocated}{127}
\newcommand{\PrecBest}{0.95}
\newcommand{\PrecBestName}{oil spill}
\newcommand{\PrecWorst}{0.57}
\newcommand{\PrecWorstName}{wildfire}
\newcommand{\ReportsMedian}{37}
\newcommand{\CoverOne}{2432}
\newcommand{\CoverMax}{719}
\newcommand{\CoverRatioOne}{4.8}
\newcommand{\CoverRatioMax}{2.0}
\newcommand{\NFailRoutes}{5}
\newcommand{\NFailVerifier}{6}
\newcommand{\FailRoutesMostWaypoints}{12}
\newcommand{\MaxKeypoints}{15}

\newcommand{\sysname}{UAV-SwarmAgent}

\newcommand{\figureorbox}[3]{%
  \IfFileExists{#1}{\includegraphics[width=#2]{#1}}{%
    \fbox{\parbox[c][#3][c]{0.95\linewidth}{\centering\small Figure file \texttt{#1} goes here}}}}

\begin{document}

\title[\sysname]{\sysname: Scalable Multi-UAV Reconnaissance and Task Allocation
  via Code Generation and Vision-Language Reasoning}

\author{Oleg Sautenkov, Yasheerah Yaqoot, Muhammad Ahsan Mustafa, Grigorii Kalinin,
  Chih-Yung Wen\texorpdfstring{\textsuperscript{\dag}}{}, and Dzmitry Tsetserukou}
\authornote{O. Sautenkov, Y. Yaqoot, M. A. Mustafa, G. Kalinin, and D. Tsetserukou
  are with the Intelligent Space Robotics Laboratory, Skolkovo Institute of Science
  and Technology, Moscow, Russia
  (\{oleg.sautenkov, yasheerah.yaqoot, ahsan.mustafa, d.tsetserukou\}@skoltech.ru).
  \textsuperscript{\dag}C.-Y. Wen is with the AiRo Lab, Department of Aeronautical and
  Aviation Engineering, The Hong Kong Polytechnic University, Hong Kong
  (chihyung.wen@polyu.edu.hk).}
\affiliation{\institution{}\country{}}

\renewcommand{\shortauthors}{Sautenkov et al.}

\begin{abstract}
Supervising a multi-UAV reconnaissance mission usually means plotting waypoints
and watching every drone. We present \sysname{}, a hierarchical multi-agent
system in which an operator states the mission in one natural-language sentence
and a ground-station agent writes and executes code that turns it into per-UAV
routes. The agent grounds inspection targets in an aerial map with a
vision-language model, splits them among $N$ UAVs by balancing route length,
and dispatches worker agents that verify incidents with a compact onboard
decision model. In simulation over \NScenarios{} incident types and \NModels{}
foundation models, the swarm located the incident in \NSuccess{} of
\NMissions{} missions (\SuccessRate\%). A paired scaling study with one to
\MaxSwarm{} UAVs shows that flight time follows an Amdahl-type law,
$T(N)=\LawFixed+\LawDivisible/N$\,s: the second UAV delivers half of the
attainable saving and the speedup is bounded at \SpeedLimit$\times$. We discuss
what this predictability, and the false alarms we observed, imply for an
operator who supervises the swarm, and outline the user study needed to test it.
\end{abstract}



\begin{teaserfigure}
  \centering
  \figureorbox{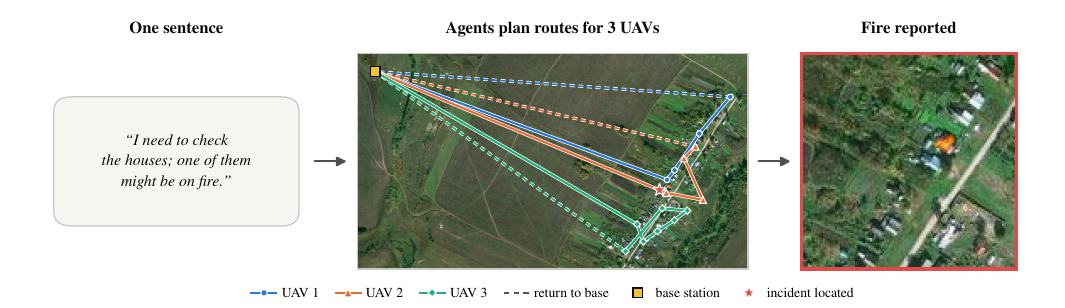}{0.9\textwidth}{1.9in}
  \caption{\sysname{} turns one operator sentence into a multi-UAV
    reconnaissance mission: the ground-station agent grounds inspection targets
    on an archival map, allocates them to the UAVs, and each UAV agent verifies
    the incident onboard and reports its location with the supporting frames.}
  \Description{Three steps from left to right, joined by arrows. First, the
    operator's sentence asking to check houses for a fire. Second, a satellite
    map on which three routes of different colours leave one base station and
    visit waypoints on the buildings of a village; a star marks where the fire
    was located. Third, the camera frame that a UAV reported, showing a
    burning house.}
  \label{fig:teaser}
\end{teaserfigure}

\maketitle

\section{Introduction}
\label{sec:intro}

\begin{figure*}[t]
  \centering
  \figureorbox{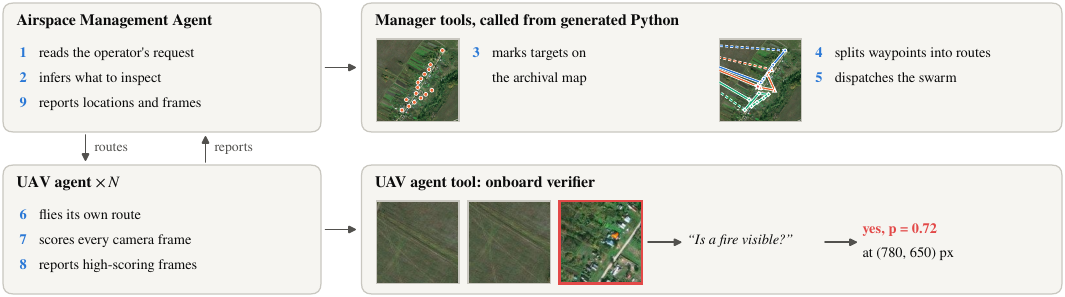}{0.88\textwidth}{1.8in}
  \caption{The two agent tiers (left), their tools (right), and the nine
    steps of a mission, with outputs of the structure-fire example of
    Figure~\ref{fig:teaser}.}
  \Description{Four boxes in two rows. Top left, the Airspace Management
    Agent: it reads the operator's request, infers what to inspect, and at the
    end reports locations and frames. Top right, its tools, called from
    generated Python: a map thumbnail with waypoints on buildings (marks
    targets on the archival map) and a map thumbnail with three routes (splits
    waypoints into routes, dispatches the swarm). Bottom left, the UAV agent,
    one per drone: it flies its own route, scores every camera frame, and
    reports high-scoring frames. Bottom right, its tool, the onboard verifier:
    three camera frames, the last showing a burning house, lead to the question
    whether a fire is visible and to the answer yes with probability 0.72 at
    map position 780, 650. Arrows link each agent to its tools; routes go down
    from the manager to the UAV agents and reports go back up.}
  \label{fig:process}
\end{figure*}

Unmanned aerial vehicles (UAVs) are routinely used for time-critical
reconnaissance such as wildfire assessment, search and rescue, and
environmental monitoring~\cite{ai_meets_uav,uav_survey}. Adding drones
shortens the mission, but it also multiplies the work of the person in charge:
each UAV needs waypoints, and each video stream needs attention. Research on
supervisory control has long shown that the number of robots one operator can
handle is limited by how long a robot can be neglected relative to how long it
takes to service it~\cite{olsen2004metrics,crandall2005validating}, and that
workload and situation awareness degrade as teams
grow~\cite{chen2014human,kolling2016human}.

Large language and vision-language models offer a different division of labour.
They can read a satellite image, follow an underspecified request, and write
programs that call tools~\cite{yao2023react,liang2023code,wang2024codeact}.
Recent aerial systems use them to navigate from language
instructions~\cite{liu2023aerialvlnvisionandlanguagenavigationuavs,lee2024citynavlanguagegoalaerialnavigation}
or to generate a single-drone mission from a
map~\cite{sautenkov2025uavvlavisionlanguageactionlargescale,sautenkov2025uavcodeagents}. What is missing
for human-swarm interaction is the step from one instruction to a
\emph{multi-drone} mission whose duration the operator can anticipate.

This late-breaking report presents \sysname{}, a system in which the operator
writes one sentence (``\emph{I need to check the houses; one of them might be
on fire}'') and receives the locations of detected incidents with the
supporting camera frames (Figure~\ref{fig:teaser}). Our contributions are:
\begin{itemize}
  \item a two-tier agent architecture in which a ground-station agent plans by
        writing code and delegates incident verification to one agent per UAV;
  \item a simulation study of \NMissions{} missions over \NScenarios{} incident
        types and \NModels{} foundation models, with mission-level success and
        frame-level false alarms reported separately;
  \item a paired scaling study with one to \MaxSwarm{} UAVs that yields a
        simple law for mission time, which an operator can use to decide how
        many drones to send.
\end{itemize}
We did not involve human participants; the implications for operators in
Section~\ref{sec:discussion} are hypotheses for the study we outline.

\section{Related Work}
\label{sec:related}

\textbf{Human-swarm interaction.} Beyond the supervisory-control limits cited
above, language has been proposed as a way to command groups at the level of
intent, for example for formation flight and
choreography~\cite{jiao2023swarmgptcombininglargelanguage,lykov2024flockgptguidinguavflocking}.

\textbf{Vision-language models for UAVs.} Benchmarks for aerial
vision-and-language navigation evaluate a single drone that follows
instructions in a simulator~\cite{liu2023aerialvlnvisionandlanguagenavigationuavs,lee2024citynavlanguagegoalaerialnavigation,EmbodiedCityzhang2024,yao2025aeroverseuavagentbenchmarksuite}.
Mission-generation systems ground language goals on satellite imagery and plan
a route for one
UAV~\cite{sautenkov2025uavvlavisionlanguageactionlargescale,sautenkov2025uavvlpavisionlanguagepathactionoptimalroute}.
UAV-CodeAgents~\cite{sautenkov2025uavcodeagents} plans such missions with
ReAct agents that ground targets by pixel pointing; \sysname{} builds on that
design and adds route allocation across $N$ drones and an onboard verifier.
Training-free multi-agent pipelines improve aerial image understanding by
routing sub-tasks to specialised tools~\cite{zhang2026advancingmllmbaseduavimage}.

\textbf{Agents that act by writing code.} ReAct interleaves reasoning with
tool calls~\cite{yao2023react}; expressing actions as executable code lets one
step compose several tools~\cite{liang2023code,wang2024codeact}. \sysname{}
applies this paradigm to swarm reconnaissance and, unlike the work above,
measures how mission time changes with the number of drones.

\section{System}
\label{sec:system}

\subsection{Agents}
\sysname{} is built on the open-source \texttt{smolagents} library
and has two tiers (Figure~\ref{fig:process}). The \emph{Airspace Management
Agent} (AMA) runs at the ground station and works in a ReAct loop whose
actions are short Python programs executed in a sandbox: it infers which
structures or areas are worth inspecting, grounds them as waypoints on the
archival map, shows them to the operator, and dispatches the swarm. One
\emph{UAV agent} per drone runs the onboard verifier on the camera frames
collected along its route.

We keep geometry out of the language model's hands. Route allocation, flight
simulation, and delegation are a single tool call, \texttt{dispatch\_swarm},
rather than code the model writes itself. In a prototype where the model
wrote this code, one model silently rescaled the waypoint coordinates before
dispatch, which shifted every route. Fixed tools
make the behaviour of the swarm independent of which model plans the mission.

\subsection{Grounding inspection targets}
\begin{figure*}[t]
  \centering
  \figureorbox{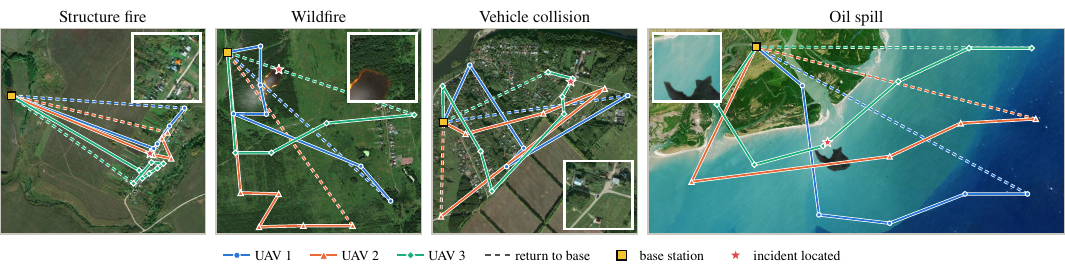}{0.95\textwidth}{1.8in}
  \caption{One example mission per scenario ($N=3$). Buildings and roads are
    grounded as \emph{objects}, one waypoint each; forest and water are
    grounded as \emph{areas} and covered with an even spread of waypoints.
    Insets show the reported frame with the highest score; dashed lines are
    the return legs to the base station. The collision and oil-spill examples
    use the refined verifier questions.}
  \Description{Four satellite maps side by side: a village with one burning
    house, a forest with a wildfire, a village with vehicles stopped at a road
    junction, and a river delta with a dark oil slick on the water. On each
    map three routes in different colours start from a yellow square base
    station and visit waypoints. On the village maps the waypoints sit on
    buildings and roads; on the forest and water maps they are spread evenly.
    A red star marks where the incident was located, and an inset shows the
    camera frame of the incident.}
  \label{fig:scenarios}
\end{figure*}

The AMA sends the map, letterboxed to a $1000\times1000$ grid, to a
vision-language model and asks for waypoints. The prompt distinguishes two
kinds of targets. \emph{Objects} such as buildings or vehicles receive one
point at their centre. \emph{Areas} such as open water or forest receive an
even spread of points, because one point in the middle of a lake does not
make a drone survey it (Figure~\ref{fig:scenarios}). Returned points are
mapped back to map pixels; at most 15 are kept. Answers in which several object points lie on one image axis at
equal spacing are discarded as fabricated.

\subsection{Allocating routes}
Waypoints are assigned to $N$ UAVs that start and land at one base station.
Because the mission ends when the last drone returns, we balance the longest
route rather than the total distance: waypoints are taken farthest-first and
each goes to the UAV whose closed route stays shortest after receiving it; each
UAV then visits its waypoints in nearest-neighbour order. Mission flight time
is the makespan
\begin{equation}
  T_{\mathrm{flight}}=\max_{k\in\{1,\dots,N\}}\frac{g\,L_k}{v},
  \label{eq:makespan}
\end{equation}
where $L_k$ is the length of route $k$ in pixels, $g=1.35$\,m/px is the ground
sampling distance of the maps, and $v=6$\,m/s is the cruise speed.

\subsection{Verifying incidents onboard}
Each UAV carries a nadir camera with a $128\times128$\,px footprint (about
170\,m on the ground). A frame is scored every 16\,m along the route by
Valen-Preview-0923, a 2B-parameter vision-language decision model that returns
the probability of ``yes'' for a scenario-specific question (for example,
whether flames or smoke are visible) in a single forward pass, without
generating text. At this size the verifier can run locally on the drone, on an
embedded NVIDIA Jetson computer; in our experiments it ran on a laptop GPU. A frame is reported when
the probability exceeds a threshold $\tau$ set per scenario (0.55--0.74). Frames of all UAVs are
batched on one GPU. Reported frames return to the AMA with their coordinates.

\section{Evaluation Setup}
\label{sec:setup}

\begin{figure}[t]
  \centering
  \figureorbox{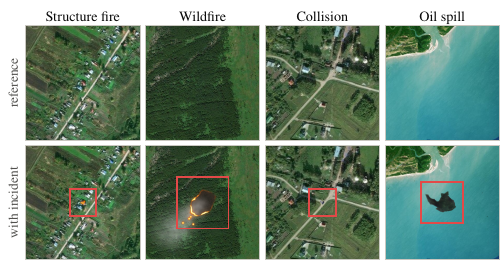}{\columnwidth}{1.7in}
  \caption{Benchmark maps: the archival reference map used for grounding
    (top) and the target map of the same area with a synthetic incident, over
    which the drones fly (bottom; incident boxed).}
  \Description{Eight satellite image crops in two rows and four columns. The
    columns show a structure fire, a wildfire, a vehicle collision, and an oil
    spill. The top row shows each area without the incident; the bottom row
    shows the same area with the incident marked by a red box: a burning house
    in a village, a burn scar with flames and smoke in a forest, vehicles
    stopped at a road junction, and a dark slick on coastal water.}
  \label{fig:dataset}
\end{figure}

\textbf{Scenarios and maps.} We use satellite maps of rural settlements derived
from the UAV-VLA
benchmark~\cite{sautenkov2025uavvlavisionlanguageactionlargescale}, each
covering about $1.4\times1.4$\,km, and five coastal scenes collected from
public web map imagery. Every mission pairs an archival \emph{reference} map, on which
targets are grounded, with a \emph{target} map of the same area that contains
the incident and supplies the camera frames
(Figure~\ref{fig:dataset}); grounding never uses the target map. Incidents are synthetic and
composited into the target map: flames on one building (structure fire, 30
maps), a burn scar with a flame front and a smoke plume inside a forest
(wildfire, 5 maps), vehicles stopped at a junction (vehicle collision, 5 maps),
and a dark slick on coastal water (oil spill, 5 maps). The operator request is
one fixed sentence per scenario and never names a location.

{\sloppy\textbf{Models.} The AMA and the UAV agents use one of \NModels{} models
through an OpenAI-compatible gateway: \texttt{gemini-3.8-flash},
\texttt{gpt-6.1-sol}, or \texttt{qwen3.8-max}. The onboard verifier is the same
in all runs and executes locally on a laptop GPU (NVIDIA RTX~4050, 6\,GB).\par}

\textbf{Protocol.} The \emph{scaling study} runs the structure-fire scenario
on all 30 maps with $N=1,\dots,\MaxSwarm$ UAVs using \texttt{gemini-3.8-flash}
(\NScalingMissions{} missions). The design is paired: every swarm size sees the
same maps and the same base station, placed at a random point of the map
border per map. The \emph{cross-model study} fixes $N=3$ and runs every
scenario with every model (\NCompared{} missions, of which the 30 Gemini fire
missions are shared with the scaling study). In total we report
\NMissions{} missions. For the collision and oil-spill scenarios the verifier
questions were later refined and the Gemini missions repeated;
Table~\ref{tab:results} gives the precision of that repeat and the mission
counts of the first runs.

\textbf{Measures.} A mission is \emph{successful} when at least one reported
frame shows the incident, that is, the incident centre lies inside the frame's
$128\times128$\,px camera footprint. Wildfires and oil spills are larger than
the footprint, so for them the frame only has to overlap the area changed by
the incident. \emph{Precision} is the share of a mission's reported frames that
satisfy the same criterion. \emph{Flight time} is the simulated makespan of
Eq.~\eqref{eq:makespan}. \emph{Agent time} is the wall-clock time of planning,
simulation, and onboard inference. \emph{Total time} is their sum, a
conservative figure because inference would overlap with flight on a real
drone.

\textbf{Analysis.} Speedups are geometric means of per-map ratios
$T(1)/T(N)$. Confidence intervals are 95\% percentile intervals from 10{,}000
bootstrap resamples of maps~\cite{efron1994bootstrap}. The effect of swarm size
is tested with Friedman's test and Page's test for ordered alternatives, with
maps as blocks. Scaling laws are fitted by least squares on log time with a
separate level per map.

\textbf{Use of generative AI.} An AI coding assistant (Claude) helped write
parts of the software, the analysis scripts, and the plotting code; the
authors checked all results.

\section{Results}
\label{sec:results}

\begin{table}[t]
  \caption{Missions in which the swarm located the incident, by scenario and
    model ($N=3$ UAVs). Prec.: median share of reported frames that show the
    incident.}
  \label{tab:results}
  \small\setlength{\tabcolsep}{4pt}
\begin{tabular}{lccccc}
\toprule
Scenario & Maps & Gemini & GPT & Qwen & Prec. \\
\midrule
Structure fire & 30 & 28/30 & 30/30 & 28/30 & 0.68 \\
Wildfire & 5 & 5/5 & 5/5 & 5/5 & 0.57 \\
Vehicle collision & 5 & 3/5 & 4/5 & 4/5 & 0.74 \\
Oil spill & 5 & 5/5 & 5/5 & 5/5 & 0.95 \\
\midrule
All scenarios & & 41/45 & 44/45 & 42/45 & \\
Agent time, median (s) & & 152 & 180 & 135 & \\
\bottomrule
\end{tabular}

\end{table}

\subsection{Mission success}
The swarm located the incident in \NSuccess{} of \NMissions{} missions
(\SuccessRate\%). Table~\ref{tab:results} breaks down the cross-model study:
all \NModels{} models completed the planning workflow in every mission, and
\NComparedLocated{} of \NCompared{} missions were successful. The scaling
study adds \NScalingLocated{} successes in \NScalingMissions{} missions. The
models differed more in latency than in outcome (last row of
Table~\ref{tab:results}). With five maps per non-fire scenario, differences
between models in those rows are within what chance alone produces.

\subsection{Scaling with swarm size}

\begin{figure}[t]
  \centering
  \includegraphics[width=\columnwidth]{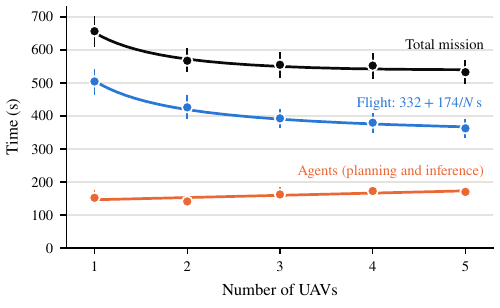}
  \caption{Scaling study (structure fire, 30 maps per swarm size): mean
    flight, agent, and total mission time with 95\% bootstrap confidence
    intervals; lines are the fitted laws.}
  \Description{A line chart of time in seconds against the number of UAVs
    from one to five. Flight time falls from about 505 to 363 seconds along a
    curve that flattens, agent time rises slightly from about 150 to 170
    seconds along a straight line, and total mission time falls from about 657
    to 533 seconds.}
  \label{fig:scaling}
\end{figure}

Flight time decreased monotonically from \FlightOne\,s with one UAV to
\FlightFive\,s with \MaxSwarm{} (\FlightGain\%; Friedman and Page tests,
$p<.001$). As a reference, photographing each map completely in parallel
strips one footprint wide would take \CoverOne\,s and \CoverMax\,s,
\CoverRatioOne{} and \CoverRatioMax{} times longer; targeting buys this time
at the risk of missing an incident that no waypoint is near. Figure~\ref{fig:scaling} shows that the gain flattens quickly. An
Amdahl-type law~\cite{amdahl1967validity}, in which a share $p$ of the
single-UAV flight time is divided among the drones and the rest is not,
\begin{equation}
  T_{\mathrm{flight}}(N)=T_{\mathrm{flight}}(1)\left[(1-p)+\frac{p}{N}\right]
  \approx \LawFixed+\frac{\LawDivisible}{N}\ \text{s},
  \label{eq:law}
\end{equation}
fits with $p=\LawP$ (95\% CI [\LawPLow, \LawPHigh]). The fixed part has a
physical reading: every drone must fly from the base station to the area of
interest and back, however the waypoints are divided. The law bounds the
flight-time speedup at $1/(1-p)=\SpeedLimit$ (CI [\SpeedLimitLow,
\SpeedLimitHigh]); the second UAV already delivers \SecondUavShare\% of the
attainable saving and \MaxSwarm{} UAVs deliver \MaxSwarmShare\%. A power law
$T\propto N^{-\PowerAlpha}$ fits equally well over this range (AIC difference
\AicGap); we prefer Eq.~\eqref{eq:law} because its parameters are
interpretable.

Agent time grew slowly with swarm size, by \AgentSlope\,s per UAV (CI
[\AgentSlopeLow, \AgentSlopeHigh]) from an intercept of \AgentIntercept\,s.
Total mission time therefore fell from \TotalOne\,s to \TotalFive\,s
(\TotalGain\%), and almost all of that gain came from the second UAV: beyond two
UAVs, consecutive swarm sizes do not differ significantly in total time
(Wilcoxon signed-rank tests, Holm-corrected).

\subsection{False alarms and missed incidents}
\label{sec:falsealarms}
Mission-level success hides what the operator would actually receive. A
mission produced a median of \ReportsMedian{} reported frames, and not all of
them show the incident: median precision ranged from \PrecWorst{} for the
\PrecWorstName{} scenario to \PrecBest{} for the \PrecBestName{} scenario
(Table~\ref{tab:results}). Of the \NFail{} failed missions, \NFailRoutes{}
failed at the planning stage: the vision-language model placed no
waypoint near the incident, so no route passed over it. In the other
\NFailVerifier{} a drone flew over the incident but the verifier did not
report it. The missions of the first kind used at most
\FailRoutesMostWaypoints{} of the \MaxKeypoints{} waypoints allowed, so
asking the model for more waypoints is a direct remedy.

\section{Discussion}
\label{sec:discussion}

\textbf{From plotting to checking.} In \sysname{} the operator's input does
not grow with the number of drones: one sentence produced routes for up to
\MaxSwarm{} UAVs. What remains for the human is verification. The waypoint
overlay shown before dispatch and the reported frames returned afterwards are
the two places where an operator can catch an error. Whether operators catch such errors, and at
what workload, is the first question for a user study.

\textbf{Swarm size as a predictable choice.} Eq.~\eqref{eq:law} shows that
in base-station reconnaissance more drones do not help proportionally: the
commute to the area does not parallelise, so a second drone is well spent
and a fifth buys little. Because the law has two
parameters with physical meaning, it could be shown to the operator as a
forecast (``two UAVs: about \FlightTwo\,s; three: about \FlightThree\,s'')
before launch. We expect such a forecast to support calibrated reliance better
than an unexplained recommendation, which remains to be tested.

\textbf{False alarms are an interaction problem.} A system that locates the
incident in most missions can still hand the operator dozens of frames to
inspect. Ranking reports by score, clustering neighbouring frames into one
report, and letting the operator adjust $\tau$ are interface decisions that
matter as much as detector accuracy.

\textbf{Limitations.} The study is a simulation with synthetic incidents and
a kinematic flight model without wind, batteries, or obstacles. Verifier
questions and thresholds were chosen on the benchmark maps, with no held-out
set. Each configuration was run once, although language-model outputs vary
between runs, and operator requests were fixed sentences. No human
participants were involved, so every statement about operators is a
hypothesis.

\section{Conclusion and Next Steps}
\label{sec:conclusion}

\sysname{} turns one natural-language instruction into a multi-UAV
reconnaissance mission. In \NMissions{} simulated missions the swarm located
the incident \SuccessRate\% of the time with \NModels{} different foundation
models, and flight time followed a simple law that makes the benefit of each
additional drone predictable. Next, we will compare instruction-based
commanding with manual waypoint planning in a user study that measures
planning time, workload, and error detection, and we will move from simulation
to physical multirotors.

\bibliographystyle{ACM-Reference-Format}
\bibliography{ref,ref_additions}

\end{document}